\documentclass[10pt,journal,compsoc]{IEEEtran}

\usepackage[nocompress]{cite}

\usepackage{url}
\usepackage{ragged2e}
\usepackage{epsfig}
\usepackage{graphicx}
\usepackage{amsmath,amssymb} 
\usepackage{mathptmx}      
\usepackage{subfigure}
\usepackage{algorithm}
\usepackage{algorithmicx}
\usepackage{algpseudocode}
\usepackage{graphics}
\usepackage{mathrsfs}
\usepackage{threeparttable}
\usepackage{color}
\usepackage[normalem]{ulem}
\usepackage{multirow}
\usepackage{float}
\usepackage{amsfonts}
\usepackage{bm}
\usepackage{array}
\usepackage[table]{xcolor}
\usepackage{colortbl}
\usepackage{pifont}
\usepackage{diagbox}
\usepackage{rotating}
\usepackage{booktabs}
\usepackage{overpic}
\usepackage{textcomp}
\usepackage{contour}
\usepackage{enumitem}
\usepackage{stfloats}
\usepackage{threeparttable}
\usepackage{makecell}
\usepackage[colorlinks=true,breaklinks=true,urlcolor=magenta]{hyperref}

\def\eg{\emph{e.g.}}
\def\etc{\emph{etc}}

\newcommand{\tabref}[1]{Table~\ref{#1}}

\newcommand{\secref}[1]{\S\ref{#1}}

\definecolor{mygray1}{gray}{.75}

\def\eg{\textit{e.g.}}

\graphicspath{{./figs/}}
\DeclareGraphicsExtensions{.jpg,.pdf,.png}

\RequirePackage{silence}
\begin{document}

\title{When SAM Meets Shadow Detection}

\author{
Leiping Jie, Hui Zhang\\
\IEEEcompsocitemizethanks{
\IEEEcompsocthanksitem Leiping Jie is with the Department of Computer Science, Hong Kong Baptist University, Hong Kong SAR, China (cslpjie@comp.hkbu.edu.hk).
\IEEEcompsocthanksitem Hui Zhang is with the Department of Computer Science and Technology, BNU-HKBU United International College, Zhuhai, China.
}
}


\IEEEtitleabstractindextext{%
\begin{abstract} \justifying
As a promptable generic object segmentation model, segment anything model (SAM) has recently attracted significant attention, and also demonstrates its powerful performance. Nevertheless, it still meets its Waterloo when encountering several tasks, \eg, medical image segmentation, camouflaged object detection, \etc. In this report, we try SAM on an unexplored popular task: shadow detection. Specifically, four benchmarks were chosen and evaluated with widely used metrics. The experimental results show that the performance for shadow detection using SAM is not satisfactory, especially when comparing with the elaborate models. Code is available at \url{https://github.com/LeipingJie/SAMShadow}.
\end{abstract}

}

\maketitle

\IEEEdisplaynontitleabstractindextext

\IEEEpeerreviewmaketitle

\IEEEraisesectionheading{\section{Introduction}\label{sec:introduction}}

\IEEEPARstart{U}{niversal} artificial intelligence (UAI) is a long pursued goal. Since the booming of deep learning, different models have been elaborately designed for various problems, even for different types of one task, or for different categories of data. Take image segmentation as an example, models designed for camouflaged object detection failed to segment lesions from medical images, or detect shadows from natural images. Recently, large language models (LLMs), \eg, ChatGPT\footnote{\url{https://chat.openai.com}}, LLaMA\footnote{\url{https://ai.facebook.com/blog/large-language-model-llama-meta-ai/}}, brought revolution to establish fundamental models. LLM were integrated into many products to facilitate the efficiency, \eg, Microsoft New Bing\footnote{\url{https://www.bing.com/new}}, Midjourney\footnote{\url{https://www.midjourney.com/}}. The success of LLMs undoubtedly made a big step towards UAI. 

Accordingly, generic purposed models for computer vision tasks are also investigated. More recently, Meta AI's SAM~\cite{kirillov2023segment} was introduced, aiming at generic image segmentation. SAM is a VIT-based model, which consists of three parts, an image encoder, a prompt encoder and a mask decoder. Two sets of prompts are supported: sparse (points, boxes, text) and dense (masks). SAM do achieve a significant performance on various segmentation benchmarks, especially its remarkable zero-shot transfer capabilities on 23 diverse segmentation datasets. However, researchers also reported its deficiencies on several tasks, such as medical image segmentation~\cite{ma2023segment, ji2023sam}, camouflaged object detection~\cite{tang2023can, ji2023sam}. It seems that the desired generic object segmentation model is still not yet implemented.

In this paper, we further try SAM on an unexplored segmentation task: shadow detection, to further evaluate the generalization ability of SAM. It is worth noting that shadow is a common phenomenon in natural images. In contrast to medical images and camouflaged images, which may occupy only a very small proportion, or even does not exist in the large training dataset of SAM (SA-1B), shadows should be more common. In particular, we compare SAM with state-of-the-art models quantitatively on four popular shadow detection datasets: SBU, UCF, ISTD and CHUK. We find that the performance of SAM on shadow detection is not satisfactory, especially for shadows cast on complicated background, and complex shadows.

\section{Experiment}\label{sec:experiment}
In this section, we will first introduce four shadow detection benchmarks, which are widely adopted to evaluate SAM. Next, the evaluation metric (\secref{sec:evaluation_metric}) is given. Considering that SAM will automatically produce multiple mask outputs for each image, we will describe our mask selection strategy (\secref{sec:mask_selection_strategy}). Then we show some predicted masks and the distribution of the number of predicted masks (\secref{sec:masks_and_distribtuion}). Finally, we present the comparison of SAM with the state-of-the-art shadow detection approaches (\secref{sec:quantitative_eval}), and provide qualitative visualizations on these datasets (\secref{sec:qualitative_eval}).

\subsection{Datasets}\label{sec:datasets}
Four widely used benchmark datasets are adopted in our experiment.
\begin{itemize}
  \item \textbf{SBU}~\cite{sbu_dataset} provides $4,089$ pairs of the shadow image and shadow map image for training, along with another $638$ pairs for testing.
  \item \textbf{UCF}~\cite{ucf_dataset} is much smaller compared with SBU, with only $145$ and $76$ training and testing pairs.
  \item \textbf{ISTD}~\cite{istd_dataset} is designed for both shadow detection and shadow removal, which means that triplets of shadow images, shadow maps and shadow-free images are provided. It uses $1,330$ of such triplets for training and $540$ for testing.
  \item \textbf{CUHK}~\cite{gy_tip2021} is the largest dataset for shadow detection. It has $7,350$ training images, $1,050$ validation images, and $2,100$ testing images in total.
\end{itemize}
Since our goal is to evaluate the performance of SAM, we only use the testing splits of these datasets.

\subsection{Evaluation Metric}\label{sec:evaluation_metric}
Balance error rate (BER) is a commonly used metric for shadow detection evaluation. It considers the performance of both shadow prediction and non-shadow prediction. BER can be formulated as:
\begin{equation}
  BER = \left(1-\frac{1}{2}\left(\frac{TP}{N_p} + \frac{TN}{N_n}\right) \right )\times 100\ ,
\end{equation}
where $TP$, $TN$, $N_p$ and $N_n$ are the number of true positive pixels, the number of true negative pixels, the number of shadow pixels and the number of non-shadow pixels respectively. For $BER$, the smaller its value, the better the performance.

\begin{table*}[htb!]
  \caption{Quantitative comparison results on SBU~\cite{sbu_dataset} dataset, UCF~\cite{ucf_dataset} and ISTD~\cite{istd_dataset}. The $\dagger$ and $\ddagger$ represent the performance evaluated with mask selection strategy \textit{F-meausure} and \textit{IOU} respectively.}
  \label{table_quantitative}
  \begin{center}
    \begin{tabular}{|c|c|ccc|ccc|ccc|}
    \hline 
    \rowcolor{mygray1}
     & & \multicolumn{3}{c|}{SBU~\cite{sbu_dataset}} & \multicolumn{3}{c|}{UCF~\cite{ucf_dataset}} & \multicolumn{3}{c|}{ISTD~\cite{istd_dataset}} \\
    \cline{3-11}
    \rowcolor{mygray1}
    \multirow{-2}{*}{Method} & \multirow{-2}{*}{Pub$_{year}$} & BER ↓ & Shadow ↓ & Non Shad.↓ & BER ↓ & Shadow ↓ & Non Shad.↓ & BER ↓ & Shadow ↓ & Non Shad.↓ \\
    \hline
    Unary-Pairwise~\cite{guo2011single} & CVPR$_{11}$ & 25.03 & 36.26 & 13.80 & - & - & - & - & - & - \\
    stacked-CNN~\cite{sbu_dataset} & ECCV$_{16}$ & 11.00 & 8.84 & 12.76 & 13.00 & 9.00 & 17.10 & 8.60 & 7.69 & 9.23 \\
    scGAN~\cite{nguyen17_iccv} & ICCV$_{17}$ & 9.10 & 8.39 & 9.69 & 11.50 & 7.74 & 15.30 & 4.70 & 3.22 & 6.18 \\
    DeshadowNet~\cite{qu17_cvpr} & CVPR$_{17}$ & 6.96 & - & - & 8.92 & - & - & - & - & - \\
    patched-CNN~\cite{hs18_iros} & IROS$_{18}$ & 11.56 & 15.60 & 7.52 & - & - & - & - & - & - \\
    ST-CGAN~\cite{istd_dataset} & CVPR$_{18}$ & 8.14 & 3.75 & 12.53 & 11.23 & 4.94 & 17.52 & 3.85 & 2.14 & 5.55 \\
    DSC~\cite{hu18_cvpr} & CVPR$_{18}$ & 5.59 & 9.76 & 1.42 & 10.54 & 18.08 & 3.00 & 3.42 & 3.85 & 3.00 \\
    ADNet~\cite{le18_eccv} & CVPR$_{18}$ & 5.37 & 4.45 & 6.30 & 9.25 & 8.37 & 10.14 & - & - & - \\
    BDRAR~\cite{zhu18_eccv} & CVPR$_{18}$ & 3.64 & 3.40 & 3.89 & 7.81 & 9.69 & 5.94 & 2.69 & \textbf{0.50} & 4.87 \\
    DC-DSPF~\cite{wang18_ijcai} & CVPR$_{18}$ & 4.90 & 4.70 & 5.10 & 7.90 & 6.50 & 9.30 & - & - & - \\
    DSDNet~\cite{zheng19_cvpr} & CVPR$_{19}$ & 3.45 & 3.33 & 3.58 & 7.59 & 9.74 & 5.44 & 2.17 & 1.36 & 2.98 \\
    MTMT-Net~\cite{chen20_cvpr} & CVPR$_{20}$ & 3.15 & 3.73 & 2.57 & 7.47 & 10.31 & 4.63 & 1.72 & 1.36 & 2.08 \\
    RCMPNet~\cite{liao_mm2021} & MM$_{21}$  & 3.13 & 2.98 & 3.28 & 6.71 & 7.66 & 5.76 & 1.61 & 1.23 & 2.00 \\
    FDRNet~\cite{Zhu_2021_ICCV} & ICCV$_{21}$ & 3.04 & 2.91 & 3.18 & 7.28 & 8.31 & 6.26 & 1.55 & 1.22 & 1.88 \\
    SynShadow~\cite{inoue2020learning} & TCSVT$_{21}$ & - & - & - & - & - & - & 1.09 & 1.13 & 1.04 \\
    TransShadow~\cite{jie2022fast} & ICASSP$_{22}$ & 3.17 & 3.75 & 3.42 & 6.95 & 9.36 & 4.55 & 1.73 & 0.99 & 2.47  \\
    SDCM ~\cite{zhu_mm2022} & MM$_{22}$ & 3.02 & - & - & 6.73 & - & - & 1.41 & - & -\\
    RMLANet~\cite{jie2022rmlanet} & ICME$_{22}$ & 2.97 & 2.53 & 3.42 & 6.41 & 6.69 & 6.14 & 1.01 & 0.68 & 1.34 \\
    MGRLN-Net~\cite{jie2022mgrln} & ACCV$_{22}$ & - & - & - & - & - & - & 1.45 & 1.65 & 1.26 \\
    FCSD-Net~\cite{jose_wacv2023} & WACV$_{23}$ & 3.15 & 2.74 & 2.56 & 6.96 & 8.32 & 5.60 & 1.69 & 0.59 & 2.79 \\
    \hline 
    \hline 
    SAM $\dagger$ & - & 25.35 & 40.72 & 9.98 & 24.09 & 40.02 & 8.17 & 26.37 & 45.3 & 7.42 \\
    SAM $\ddagger$ & - & 28.82 & 50.13 & 7.51 & 30.51 & 50.13 & 7.51 & 30.51 & 54.35 & 6.67 \\
    \hline 
    \end{tabular}
    \vspace{-3em}
  \end{center}
\end{table*}

\begin{figure}[htb!]
  \centering
  \includegraphics[width=0.85\linewidth]{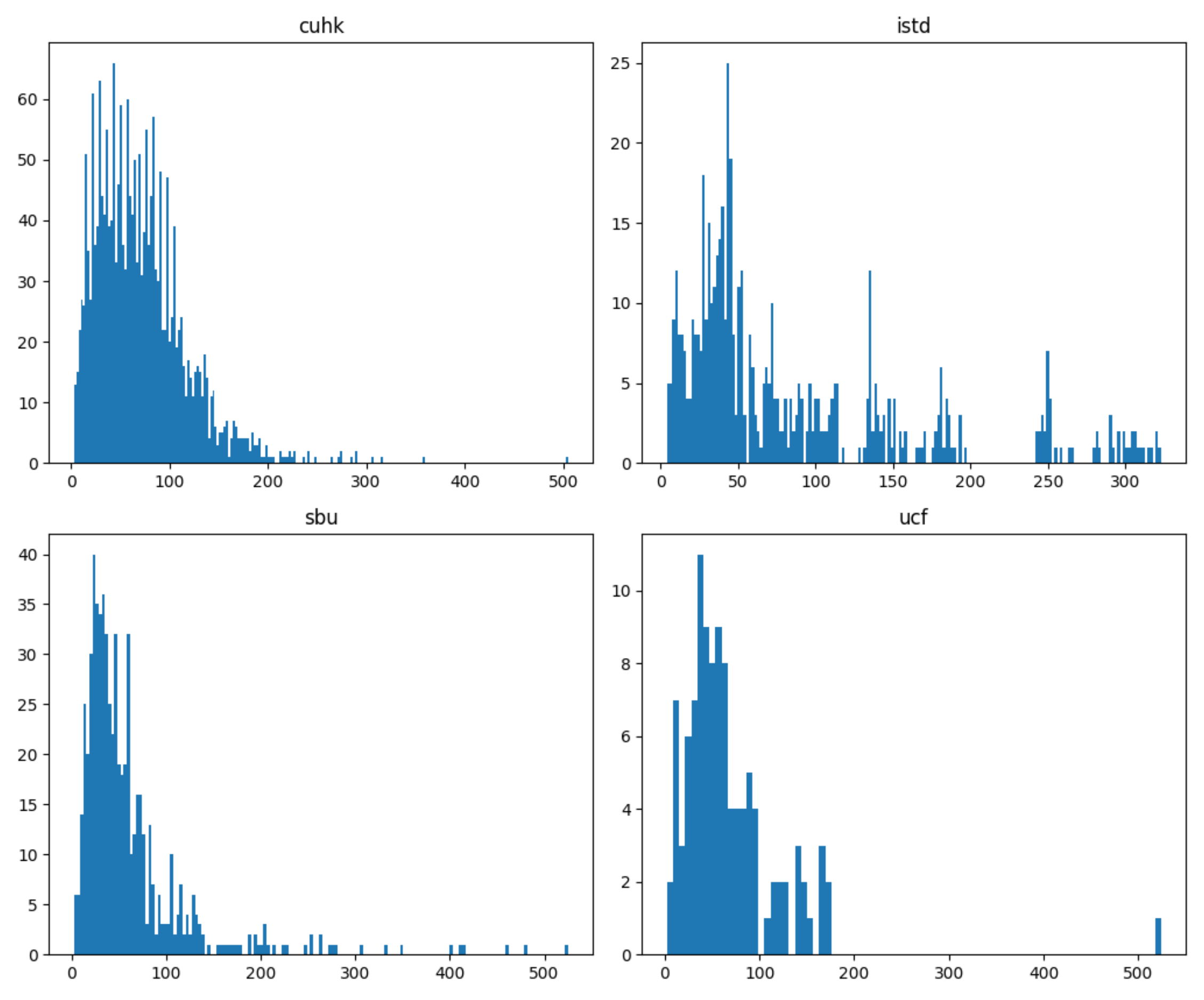}
  \caption{Histograms for visualizing the distribution of the number of predicted masks on our selected four datasets.
 }
  \label{fig:number_distribution}
\end{figure}

\begin{figure}[htb!]
  \centering
  \includegraphics[width=0.85\linewidth]{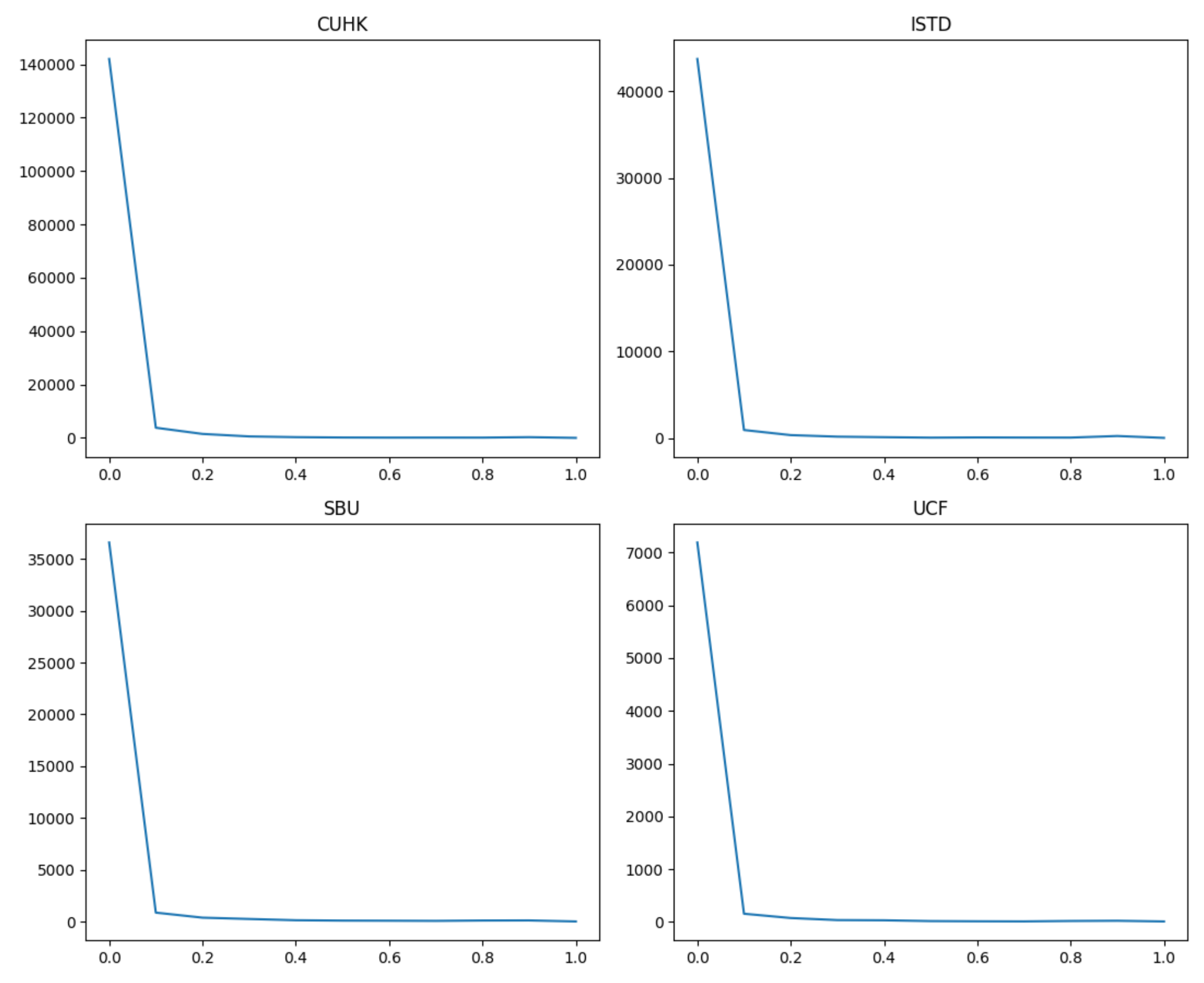}
  \caption{The mask ratio distribution on our selected four datasets. SAM tends to produce more tiny segmentation masks that only occupy less than $10\%$ of the total image. 
 }
  \label{fig:pixel_ratio}
\end{figure}

\subsection{Mask Selection Strategy}\label{sec:mask_selection_strategy}
When no prompts are fed to SAM, it will automatically generate multiple binary masks for an input image. These masks cover nearly all parts of the image. Thus, how to select the best matched candidate is crucial and may have great impact on the performance. Given an input image $I$, SAM will generate $N$ binary masks $\{O\}^N_{i=1}$. Here, we investigate two different strategies:
\begin{itemize}
  \item \textbf{Max F-measure}. This strategy was originally used in ~\cite{tang2023can}. To be specific, the mean F-measure scores $F_i$ for all predicted masks are calculated, and the max one is selected as the predicted mask $O$ for the image $I$. We formulate it as follows:
  \begin{equation}
    O = O_{\underset {i} { \operatorname {arg\,max} } \, (F_1, F_2, ..., F_i..., F_N)}. 
  \end{equation}

  \item \textbf{Max IoU}. This strategy was originally adopted in ~\cite{ji2023sam}. Specifically, the IoU between each predicted mask $O_i$ and its corresponding ground truth $G$ is calculated. Then, the mask that has the highest IoU score with $G$ is selected as the predicted mask of for image $I$. This can be formulated as:
  \begin{equation}
    O = O_{\underset {i} { \operatorname {arg\,max} } \, (IoU(O_1, G), IoU(O_2, G), ..., IoU(O_i, G)..., IoU(O_N, G))}. 
  \end{equation}
\end{itemize}

\subsection{Predicted Masks and the Distrbution}\label{sec:masks_and_distribtuion}
As mentioned ahead, the number of predicted masks varies with different input images, when no prompts are provided. We visualize three images with their predicted masks by SAM in~\ref{fig:varing_pred_masks}. To better understand how widely the range might be, the distributions of the number of predicted masks on our selected four datasets is presented in~\ref{fig:number_distribution}. The number of predicted masks varies from $0$ to even $500$ on the SBU, UCF and CUHK dataset. However, most of them are within $[0, 100]$.

\begin{figure*}[htb!]
  \centering
  \includegraphics[width=\linewidth]{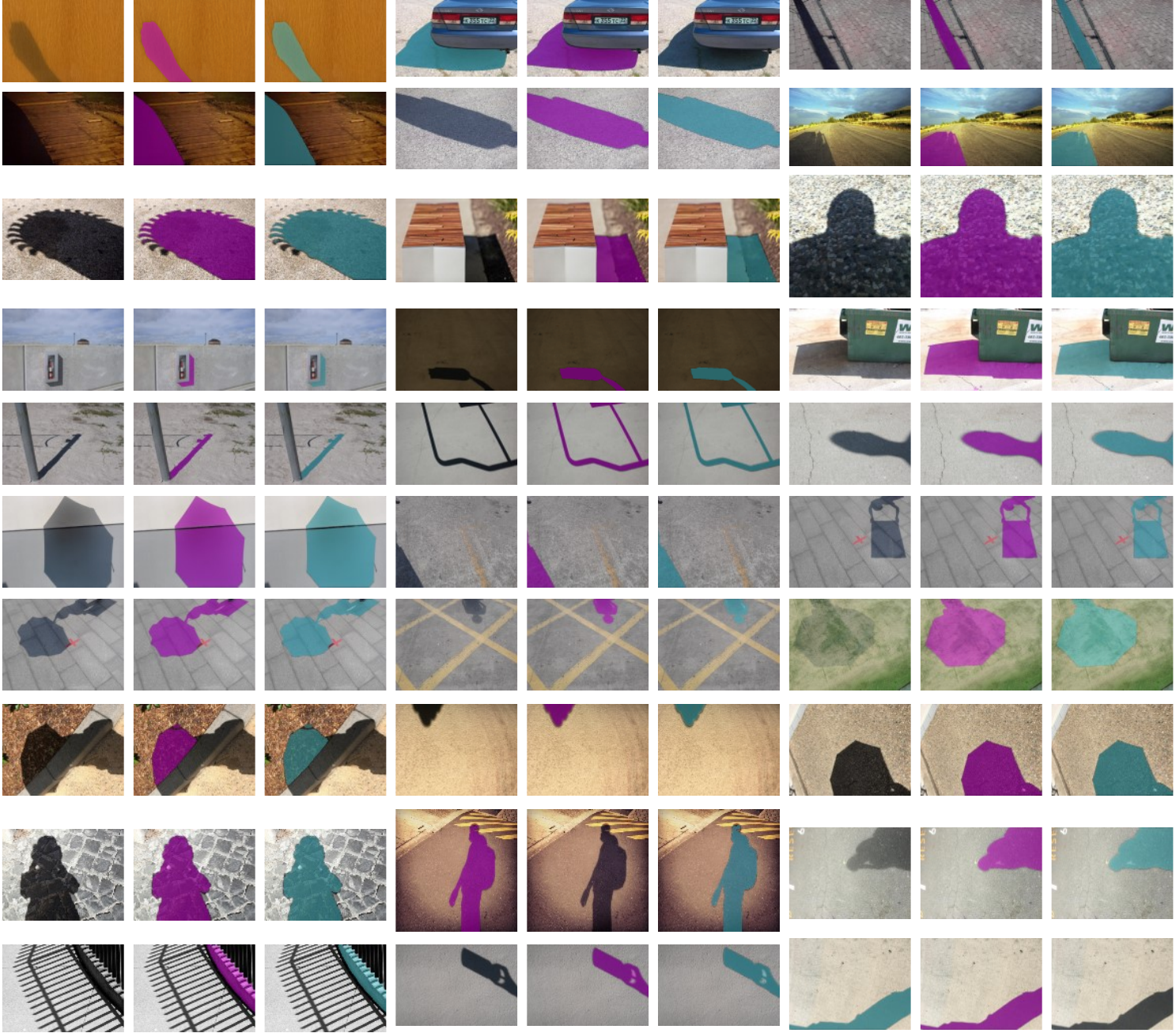}
  \caption{Visualization of good predictions. We show a triplet image: the input image, the predicted mask (\textcolor[RGB]{224, 122, 222}{purple}) using F-measure as mask selection strategy, and those (\textcolor[RGB]{155, 213, 217}{cyan}) using IOU as mask selection strategy. Specifically, the first, second and third row are selected from the SBU dataset, the 4th and 5th row are from the UCF dataset, the 6th and the 7th row are results of the ISTD dataset, while the last three rows are from the CUHK dataset.
 }
  \label{fig:qualitative_eval_sucess}
\end{figure*}

\subsection{Predicted Masks and the Distrbution}\label{sec:pixel_ratio}
\begin{figure*}[htb!]
  \centering
  \includegraphics[width=\linewidth]{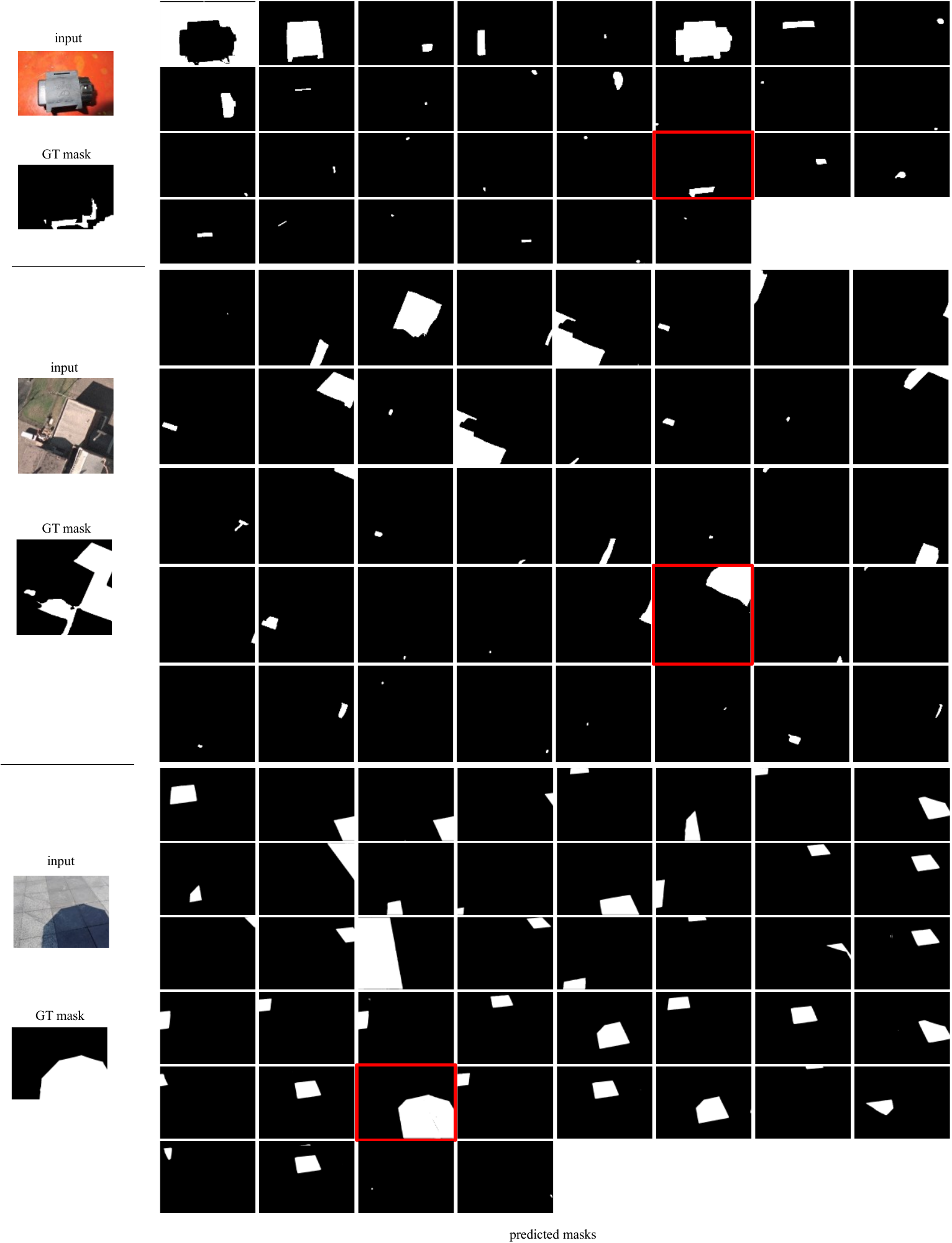}
  \vspace{-20pt}
  \caption{Visualization of SAM's predicted masks for three input images: $\$\_35.jpg$, $04822305\_2049\_2305\_2305\_2561.jpg$, $128{-}1.png$, from SBU, UCF and ISTD dataset respectively. The masks with red frame are the selected predictions using the \textit{F-measure} mask selection strategy.
 }
  \label{fig:varing_pred_masks}
\end{figure*}

We empirically found that SAM tends to produce too many tiny segmentations, as shown in~\ref{fig:varing_pred_masks}. Fig~\ref{fig:pixel_ratio} presents the mask ratio distribution on our selected four datasets. Another interesting observation is that SAM seems to segment with different thresholds. As shown in the first row of prediction masks in~\ref{fig:varing_pred_masks}, the second and the fourth predictions belong to the same object. More importantly, their appearances also are similar. Why SAM produces two different masks for the same object may be interesting to answer.

\subsection{Quantitative Evaluation}\label{sec:quantitative_eval}
The quantitative comparison is reported in~\tabref{table_quantitative} and ~\tabref{table_detection_cuhk}. Compared with existing SOTA shadow detection methods, SAM performs the worst and remains a large gap. The Ber metric for SOTA approaches is around 3 on the SBU dataset, while SAM's result is around 25, which is far from satisfactory. Similar conclusions also occur on the UCF, ISTD and CUHK dataset. Moreover, the F-measure performs better than the IOU when using as mask selection strategy.

\begin{table}[htbp]
  \begin{center}
      \caption{Quantitative comparison with the state-of-the-art methods for shadow detection on the CUHK~\cite{gy_tip2021} dataset. The $\dagger$ and $\ddagger$ represent the performance evaluated with mask selection strategy \textit{F-meausure} and \textit{IOU} respectively.}
      \begin{tabular}{|c|c|c|}
      \hline 
      \rowcolor{mygray1}
      Method & Pub$_{year}$ & BER $\downarrow$\\
      \hline 
      A+D Net~\cite{le18_eccv} & ECCV$_{18}$ & 12.43 \\
      BDRAR~\cite{zhu18_eccv} & ECCV$_{18}$ & 9.18 \\
      DSC~\cite{hu18_cvpr} & DSC$_{18}$ & 8.65 \\
      DSDNet~\cite{zheng19_cvpr} & CVPR$_{19}$ & 8.27 \\
      RCMPNet~\cite{liao_mm2021} & MM$_{21}$ & 21.23 \\
      FSDNet~\cite{gy_tip2021} & TIP$_{21}$ & 8.65 \\
      \hline
      \hline 
      SAM $\dagger$ & - & 30.23 \\
      SAM $\ddagger$ & - & 33.19 \\
      \hline 
      \end{tabular}
      \label{table_detection_cuhk}
      \end{center}
\end{table}

\begin{figure*}[htb!]
  \centering
  \includegraphics[width=\linewidth]{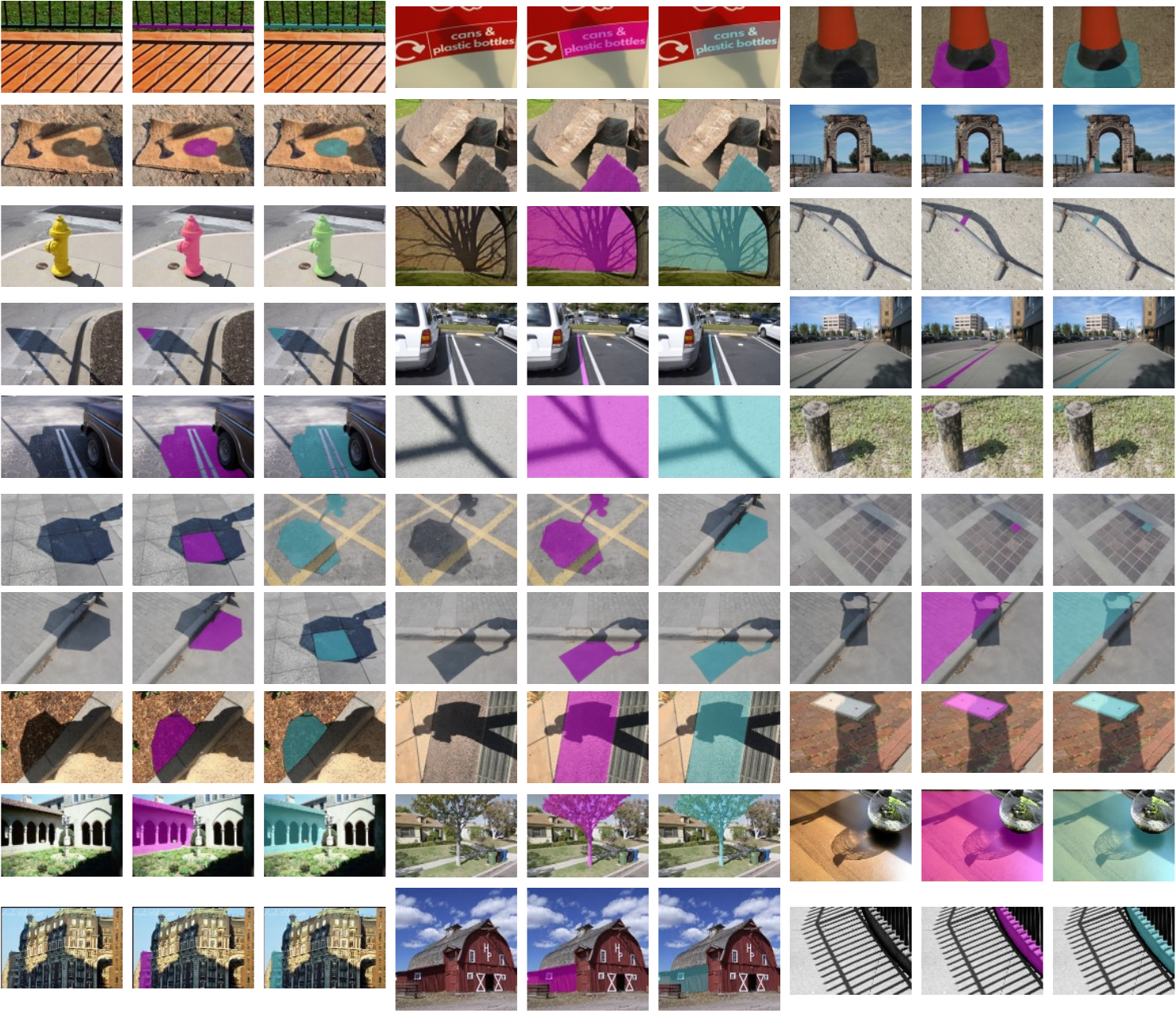}
  \caption{Visualization of failure cases. We show a triplet image: the input image, the predicted mask (\textcolor[RGB]{224, 122, 222}{purple}) using F-measure as mask selection strategy, and those (\textcolor[RGB]{155, 213, 217}{cyan}) using IOU as mask selection strategy. Specifically, the first, second and third row are selected from the SBU dataset, the 4th and 5th row are from the UCF dataset, the 6th and the 7th row are results of the ISTD dataset, while the last three rows are from the CUHK dataset.
 }
  \label{fig:qualitative_eval_fail}
\end{figure*}

\subsection{Qualitative Evaluation}\label{sec:qualitative_eval}
Fig~\ref{fig:qualitative_eval_sucess} shows some good examples. Visually, the backgrounds of these predictions are usually simple, or the shadows have a high contrast intensity. Apart from these successful cases, most of SAM's predictions are not satisfactory, as shown in Fig~\ref{fig:qualitative_eval_fail}. In particular, SAM seems to struggle with complex shadows (\eg, the middle images of the third row), or those shadows cast on complicated backgrounds (\eg, the middle images of the first row).

\subsection{Discussion}\label{sec:discussion}
Based on the experiments, we conclude that:
\begin{itemize}
  \item The performance of SAM on shadow detection is far from satisfactory.
  \item SAM tends to produce too many tiny segmentation for shadow images.
  \item SAM often fails when facing complex shadows.
  \item SAM usually struggles to segment shadows that are cast on complicated backgrounds.
\end{itemize}

However, several pioneers ~\cite{chen2023sam, wu2023medical, zhang2023customized} attempted to adopt SAM to specific datasets, and demonstrated satisfactory performance. This indicates the potential of SAM being adopted for shadow detection.

\section{Conclusion}\label{sec:conclusion}
This paper conducts preliminary experiments on shadow detection with SAM. We analyze the predicted masks using the distribution of the number of generated masks and that of the pixel ratio of each predicted masks. Quantitative and qualitative evaluations are also presented. However, we find that current SAM is not suitable with the automatic setting and needs further adaption for shadow detection problem. We expect our paper can provide valuable information for the development of applying SAM on shadow detection and other related fields.

\ifCLASSOPTIONcaptionsoff
  \newpage
\fi

{
\bibliographystyle{IEEEtran}
\bibliography{bibliography}
}

\vfill

\end{document}